\documentclass[letterpaper]{article} 
\usepackage{aaai2027}  
\usepackage[hyphens]{url}  
\usepackage{graphicx} 
\usepackage{natbib}  
\usepackage{caption} 
\usepackage{algorithm}
\usepackage{algorithmic}
\usepackage{adjustbox}
\usepackage{makecell}

\usepackage{amsmath, amsfonts}
\usepackage{longtable}

\usepackage{newfloat}
\usepackage{listings}
\DeclareCaptionStyle{ruled}{labelfont=normalfont,labelsep=colon,strut=off} 
\floatstyle{ruled}
\newfloat{listing}{tb}{lst}{}
\floatname{listing}{Listing}

\usepackage{booktabs}

\usepackage{threeparttable}
\usepackage{url}

\usepackage{makecell}

\usepackage{graphicx}
\usepackage{booktabs}
\usepackage{multirow}
\usepackage{xcolor}

\usepackage{xcolor}
\definecolor{crpscolor}{rgb}{0.85, 0, 0}
\definecolor{picppurplecolor}{HTML}{800080}
\definecolor{mpiwcolor}{HTML}{D97706}
\definecolor{miscolor}{rgb}{0, 0.2, 0.85}
\definecolor{maecolor}{rgb}{0.85, 0.25, 0.55}
\definecolor{rmsecolor}{rgb}{0.1, 0.65, 0.2}

\newcommand{\ms}[2]{%
  \substack{#1\\[0.25ex]\scriptstyle (\pm #2)}%
}

\newcommand{\crpsbest}[2]{%
  \textcolor{crpscolor}{\substack{\underline{\mathbf{#1}}\\[0.25ex]
  \scriptstyle\underline{\mathbf{(\pm #2)}}}}%
}
\newcommand{\picpbest}[2]{%
  \textcolor{picppurplecolor}{\substack{\underline{\mathbf{#1}}\\[0.25ex]
  \scriptstyle\underline{\mathbf{(\pm #2)}}}}%
}

\newcommand{\maebest}[2]{%
  \textcolor{maecolor}{\substack{\underline{\mathbf{#1}}\\[0.25ex]
  \scriptstyle\underline{\mathbf{(\pm #2)}}}}%
}
\newcommand{\rmsebest}[2]{%
  \textcolor{rmsecolor}{\substack{\underline{\mathbf{#1}}\\[0.25ex]
  \scriptstyle\underline{\mathbf{(\pm #2)}}}}%
}
\newcommand{\mpiwbest}[2]{%
  \textcolor{mpiwcolor}{\substack{\underline{\mathbf{#1}}\\[0.25ex]
  \scriptstyle\underline{\mathbf{(\pm #2)}}}}%
}

\definecolor{crpscolor}{rgb}{0.85, 0, 0}
\definecolor{picppurplecolor}{HTML}{800080}
\definecolor{mpiwcolor}{rgb}{0, 0.2, 0.85}
\definecolor{maecolor}{rgb}{0.85, 0.25, 0.55}
\definecolor{rmsecolor}{rgb}{0.1, 0.65, 0.2}

\title{EngressionTS: Turning Deep Time Series Models into Probabilistic Forecasters}

\author{
    Anusha Tomar\textsuperscript{\rm 1,3},
    Rajdeep Pathak\textsuperscript{\rm 2,3},
    Tanujit Chakraborty\textsuperscript{\rm 2,3}
}

\affiliations{
    \textsuperscript{\rm 1}Guru Gobind Singh Indraprastha University, Delhi, India\\
    \textsuperscript{\rm 2}SCAI, Sorbonne Université, Paris, France\\
    \textsuperscript{\rm 3}SAFIR, Sorbonne University Abu Dhabi, UAE\\
    anushatomar913@gmail.com, rajdeep.pathak@sorbonne.ae, tanujit.chakraborty@sorbonne.ae
}

\begin{document}

\maketitle

\begin{abstract}
Deep learning has substantially advanced time series (TS) forecasting, yet many state-of-the-art deep TS models do not naturally quantify predictive uncertainty. Existing probabilistic forecasting approaches typically address this limitation through parametric likelihoods, quantile-based objectives, or post-hoc conformal calibration. However, parametric methods rely on distributional assumptions that may be misspecified, quantile regression requires explicitly estimating selected conditional quantiles, and conformal approaches often introduce an additional calibration stage. We introduce \texttt{EngressionTS}, a general framework that integrates engression (a distributional learning method) with deep TS architectures. By formulating engression as a model-intrinsic yet architecture-agnostic framework, \texttt{EngressionTS} enables standard deep learning models to directly output probabilistic intervals. We empirically evaluate the engression-augmented models against established probabilistic techniques across three standard multivariate benchmarks. Our Python package, \texttt{EngressionTS}, facilitates seamless, plug-and-play training and inference and outputs reliable probabilistic forecasts.
\end{abstract}

\begin{links}
    \link{Code}{https://github.com/anushatomar13/engressionts}
    \link{Documentation}{https://engressionts.vercel.app/}
\end{links}

\section{Introduction}
Probabilistic forecasting is essential for characterizing uncertainty in time series analysis \citep{besher2026beyond, panja2023probabilistic}. Recently, Engression was introduced by \citet{shen2025engression} as a deep distributional regression framework that estimates the conditional distribution of a target variable given covariates by optimizing an energy score objective under an additive noise paradigm. The engression methodology has been extended to multivariate time series and spatiotemporal paradigms \citep{kraft2026engression, pathak2026entransformer, pathak2026deep} for probabilistic forecasting.
Building upon these developments, we introduce \texttt{EngressionTS}, an open-source general framework that extends existing deep TS forecasting models to produce probabilistic forecasts through stochastic predictions and energy score-based training. Rather than specifying a parametric output distribution, the model generates multiple plausible future trajectories that form an empirical predictive distribution, from which specific quantiles may be derived for uncertainty quantification. Experimental results on three multivariate benchmark datasets demonstrate highly competitive performance of engression-based deep TS models on probabilistic forecasting.

Our contributions can be summarized as follows.
\begin{itemize}
   \item We introduce \texttt{EngressionTS}, developed with an \emph{open-source-friendly approach} as a Python package for probabilistic time series forecasting.
    
    \item We integrate engression with popular neural TS architectures, establishing it as an architecture-agnostic yet model-intrinsic framework that enables deep models to directly generate probabilistic forecasts.
    
    
    \item We evaluate the performance of engression-augmented deep TS models against parametric Gaussian likelihoods (GL), quantile regression (QR), conformal prediction (CP), and conformalized quantile regression (CQR) augmented baselines on three benchmark datasets.
\end{itemize}

\section{Methodology}\label{sec:methodology}
Let $[\mathbf{y}_1, \mathbf{y}_2, \ldots, \mathbf{y}_T]^\top \in \mathbb{R}^{T\times N}$ represent a multivariate time series dataset, where $\mathbf{y}_t\in \mathbb{R}^N$ denotes the observation at time $t$, and $T$ and $N$ specify the total number of temporal steps and variables, respectively. Given a historical look-back window of length $p$, defined as $\mathbf{Y}_{t-p+1:t} = [\mathbf{y}_{t-p+1}, \ldots, \mathbf{y}_t]$, the primary objective of the forecasting task is to estimate the subsequent sequence over a specified prediction horizon $q$, expressed as $\mathbf{\widehat{Y}}_{t+1:t+q} = [\mathbf{\widehat{y}}_{t+1}, \ldots, \mathbf{\widehat{y}}_{t+q}]$ \citep{barman2025epidemic}. To facilitate generative forecasting, the engression framework employs a pre-additive noise injection mechanism directly on the model inputs. The historical input matrix is duplicated $M$ times along the batch dimension to construct a representative ensemble. An independent stochastic noise tensor $\epsilon^{(m)}$ is sampled from a prescribed probability distribution and added to the historical features prior to deep temporal processing, resulting in the perturbed input sequences $\mathbf{Y}'^{(m)}_{t-p+1:t} = \mathbf{Y}^{(m)}_{t-p+1:t} + \epsilon^{(m)}$, for $m=1,\ldots,M$. 
The chosen deep neural network is trained by optimizing the empirical energy score loss, which directly evaluates the quality of multivariate probabilistic forecasts utilizing the generated samples:
\begin{equation*}
\resizebox{\linewidth}{!}{$
    \mathcal{L}_{ES} = \frac{1}{M} \sum_{m=1}^{M} \left\| \mathbf{\widehat{Y}}^{(m)} - \mathbf{Y} \right\|_2 - \frac{1}{2M(M-1)} \sum_{i=1}^{M} \sum_{j=1}^{M} \left\| \mathbf{\widehat{Y}}^{(i)} - \mathbf{\widehat{Y}}^{(j)} \right\|_2
$}.
\end{equation*}
Once trained, the resulting \texttt{EngressionTS} architecture generates a diverse ensemble of $M$ plausible multi-step forecast trajectories, $\{\mathbf{\widehat{Y}}^{(m)}_{t+1:t+q}\}_{m=1}^M$. Distinct forecast trajectories can be sampled from the trained model via forward simulation using different noise samples to obtain probabilistic forecasts from \texttt{EngressionTS} models. This way, the idea of engression can be incorporated with deep TS models to turn them into probabilistic forecasters. 



\begin{figure}[!t]
\centering\includegraphics[width=\linewidth]{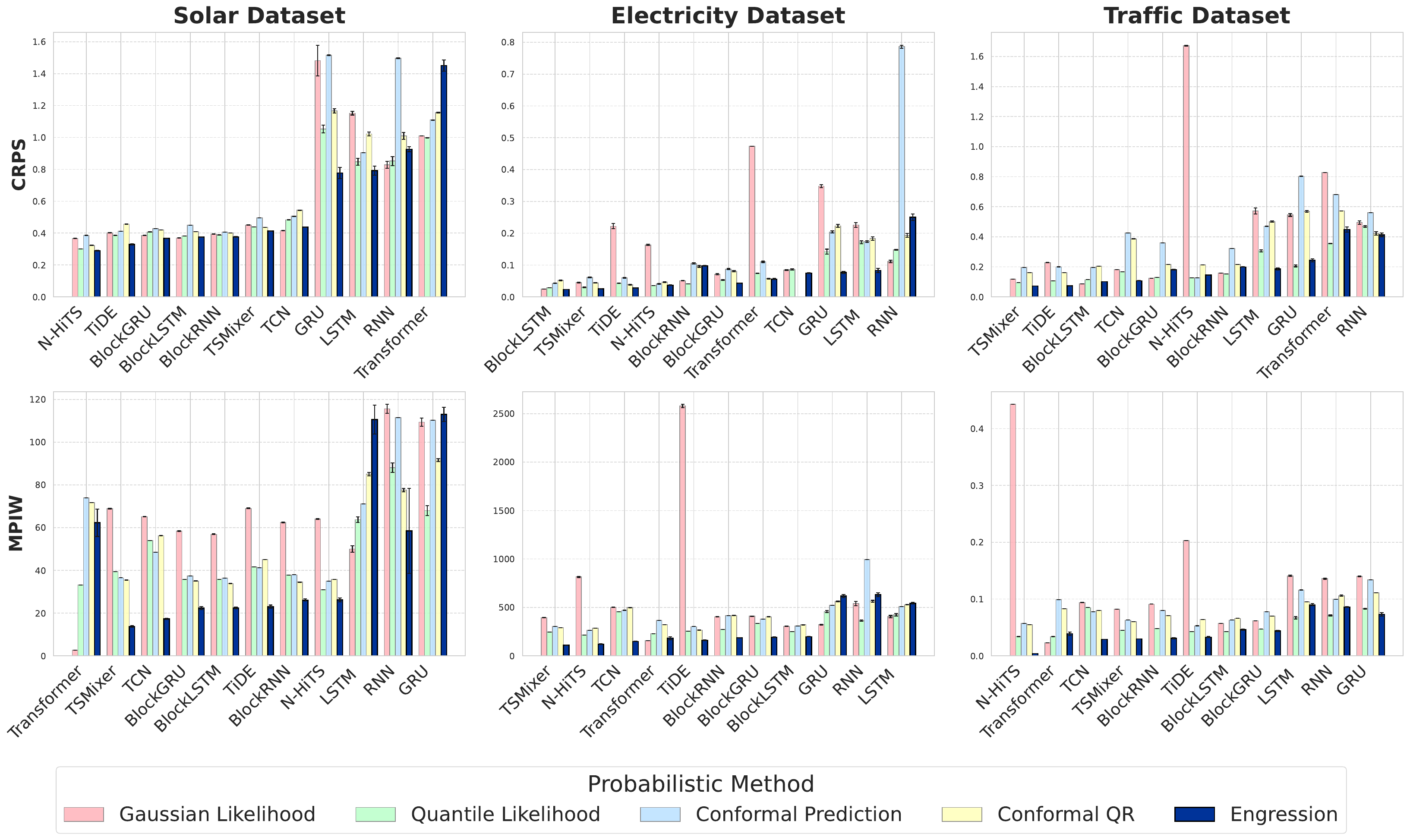}
\caption{{\small CRPS (top) and MPIW (bottom) across models and probabilistic methods for the Solar, Electricity, and Traffic datasets.}}\label{fig:crps_mpiw_results}
\end{figure}

\section{Experimental Analysis}
We evaluate \texttt{EngressionTS} on three widely-used multivariate time series forecasting benchmarks, namely Solar, Electricity, and Traffic. The Solar dataset contains solar power production data from 137 photovoltaic plants in Alabama State. The dataset features a 10-minute sampling rate over the year 2006, which is aggregated to hourly intervals, resulting in 7009 time steps in the training set. The Electricity dataset contains hourly electricity consumption across 370 clients from 2012 to 2014. Finally, the Traffic dataset consists of hourly road occupancy rates measured by 963 sensors in the San Francisco Bay Area freeway system between 2015 and 2016. We use the standard train-test split for our experiments as in GluonTS \citep{alexandrov2020gluonts}. The test set consists of 7 windows for each dataset, starting from the time-step where the training set ends. The datasets are publicly available, and their key properties are summarized in Table~\ref{tab:temporal_data_characteristics}.

\begin{figure}[!ht]
\centering\includegraphics[width=0.95\linewidth]{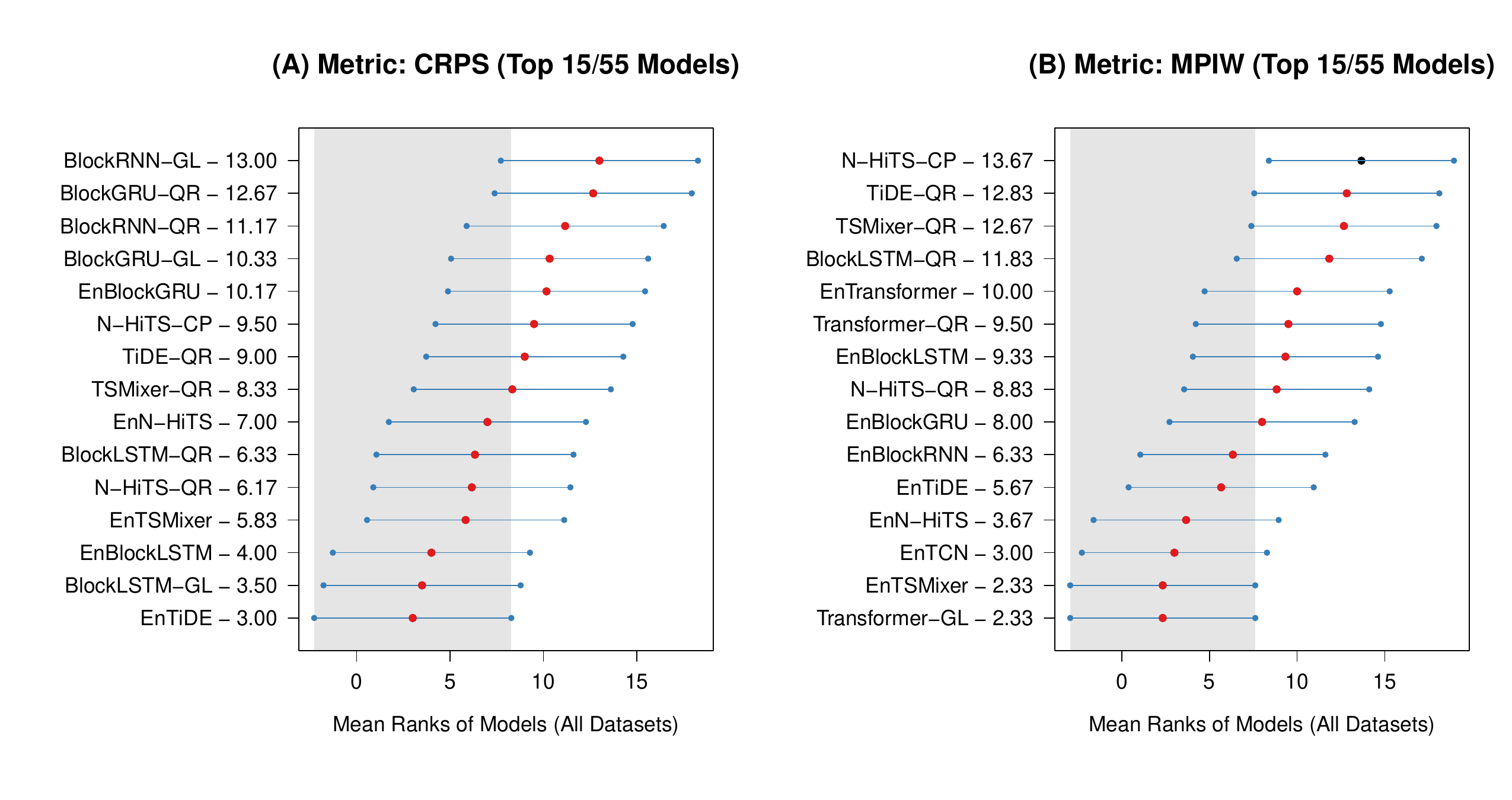}
\caption{{\small MCB test results for (A) CRPS and (B) MPIW metrics across all three datasets.}}\label{fig:mcb}
\end{figure}

Comparison is performed against GL, QP, CP, and CQR-augmented deep TS models detailed in Appendix~\ref{appendix:baselines}, with both point and probabilistic metrics listed in Appendix~\ref{appendix:metrics}. Figure~\ref{fig:crps_mpiw_results} summarizes the results obtained for two probabilistic metrics, namely CRPS and MPIW. Across the datasets, engression-based models show strong probabilistic performance (lower CRPS) with a good balance between interval coverage and sharpness (lower MPIW), as compared to GL-, QR-, and CP-based methods. Specifically, engression models dominate on the CRPS metric in $\sim 64\%$ and $\sim 73\%$ of the cases on the Solar and Electricity datasets, respectively, as compared to their probabilistic counterparts. They consistently rank among the top performing frameworks overall, as depicted by the Multiple Comparisons with the Best (MCB) test in Figure~\ref{fig:mcb}. The quantitative results, more details on the experiments, and further visualizations are provided in the appendix.

\section{Software Implementation}
We built \texttt{EngressionTS}\footnote{\url{https://pypi.org/project/engressionts/}} as a modular Python framework with separate components for the base engression class, energy score loss, stochastic noise layers, and forecasting backbones. It provides engression-based implementations of deep neural TS forecasting architectures listed in Table~\ref{tab:forecasting_models}, and follows the standard user-friendly \texttt{fit()} and \texttt{predict()} interface for training and inference, respectively. The core class, \texttt{EngressionPLModule}, builds on Darts' \texttt{PLForecastingModule} \citep{darts2020} with a PyTorch-Lightning backend and handles shared training and inference logic. Custom PyTorch forecasting models can be integrated by wrapping them with \texttt{EngressionPLModule}. 
During training, the in-sample predictions are reshaped into $(M, \text{Batch}, q, N)$ and are used to compute the energy score loss. The parameter \texttt{num\_samples\_train} determines the number of stochastic samples during training, used in the computation of the energy score. During inference, \texttt{num\_samples} determines the number of independently generated forecast trajectories, representing the empirical predictive distribution. The models produce outputs of shape $(q, N, M_{\text{infer}})$, from which specific quantiles can be derived along the last dimension to get point forecasts and prediction intervals. 


\section{Conclusion}
We presented \texttt{EngressionTS}, a general framework and a user-friendly Python software that integrates engression with popular neural TS forecasting models to make them probabilistic. We establish that engression is highly competitive and flexible as compared to existing distributional, quantile, and conformal approaches for probabilistic forecasting. The Python package enables seamless implementation of engression-augmented models, which we posit to be strong baselines for benchmarking on probabilistic forecasting tasks. Integrating other deep neural forecasting models in \texttt{EngressionTS} remains a promising future direction.

\renewcommand{\thetable}{A.\arabic{table}}
\renewcommand{\thefigure}{A.\arabic{figure}}
\setcounter{table}{0}
\setcounter{figure}{0}
\appendix
\section*{Appendix}
\section{Experimental Setup}\label{appendix:experiments}

\subsection{Data Characteristics}


\begin{table}[!h]
    \centering
    \caption{Key characteristics of the temporal benchmark datasets used in this study.}
\label{tab:temporal_data_characteristics}
\begin{adjustbox}{width=\linewidth}
    \begin{tabular}{cccccccc}
    \toprule Name & Granularity & Dim. & \makecell{\# Training\\Samples} & \makecell{Context\\Length} & \makecell{Pred.\\Length} & \makecell{Seq.\\Lags} & \makecell{\# Test\\ Windows}   \\
   
    \midrule Solar\textsuperscript{a} & 1 hour & 137 & 7009 & 24 & 24 & (1, 24, 168) & 7\\
           Electricity\textsuperscript{b} & 1 hour & 370 & 5833 & 24 & 24 & (1, 24, 168) & 7\\
           Traffic\textsuperscript{c} & 1 hour & 963 & 4001 & 24 & 24 & (1, 24, 168) & 7\\ \bottomrule
    \end{tabular}
    \end{adjustbox}
    \vspace{1mm}
    \begin{minipage}{\linewidth}
    \tiny
    \textsuperscript{a}\url{https://www.nrel.gov/grid/solar-power-data.html};
    \textsuperscript{b}\url{https://archive.ics.uci.edu/dataset/321/electricityloaddiagrams20112014};
    \textsuperscript{c}\url{https://archive.ics.uci.edu/dataset/204/pems-sf};
    \end{minipage}
\end{table}

\subsection{Baselines}\label{appendix:baselines}
Our evaluation framework compares \texttt{EngressionTS} as a method for probabilistic forecasting against four well-established classes of methods, as elaborated below.
\begin{itemize}
    \item \textbf{Parametric likelihood}: These models specify a base emission distribution and predict its parameters directly from the historical data. Classical deep neural TS models include DeepAR \citep{salinas2020deepar}, which utilizes a Gaussian likelihood in a recurrent neural network (RNN) for continuous variables and a Negative Binomial likelihood for count outcomes. Such parametric frameworks are trained to minimize the negative log-likelihood (NLL) of the chosen emission distribution. We utilize the Gaussian likelihood with neural time series models to compare against this class.
    
    \item \textbf{Quantile regression (QR)}: Quantile regression \citep{koenker1978regression} is a non-parametric method trained using the asymmetric pinball loss \citep{ingo2011pinball}. It estimates selected quantiles of the predictive distribution, without requiring any distributional assumptions. We estimate nine quantile levels $\tau \in \{0.1, \ldots, 0.9\}$ for all models in our experiments under this class. 
    
   \item \textbf{Conformal prediction (CP)}: Conformal prediction \citep{shafer2008tutorial} is a model-agnostic, post-hoc calibration technique that provides distribution-free coverage guarantees. We use the naive conformal prediction in our experiments, where absolute residual errors (non-conformity scores) are calculated on a hold-out calibration set to obtain a symmetric interval margin. 
   
    \item \textbf{Conformalized Quantile Regression (CQR)}: CQR combines the input-dependent widths of QR with the coverage guaranties of CP \citep{romano2019conformalized}. The base model estimates quantiles like standard QR, and the calibration set is used to compute adjustment scores based on the distance between the true targets and the predicted percentile bounds.
\end{itemize}


To demonstrate the architecture-agnostic capability of engression, we integrate it with a diverse set of neural forecasting models in \texttt{EngressionTS}. These models span four primary architectural categories, as detailed in Table~\ref{tab:forecasting_models}.

\begin{table*}[!ht]
\centering
\caption{Forecasting models offered in the first version of \texttt{EngressionTS}.}
\label{tab:forecasting_models}
\begin{adjustbox}{width=\linewidth}
\begin{tabular}{cccc}
\toprule
\textbf{Model} & \textbf{Base Model} & \textbf{Model Class} & \textbf{Reference} \\
\midrule
EnRNN & Vanilla RNN (Recurrent Neural Network) & Recurrent Network & \citet{rumelhart1986learning} \\
EnLSTM & LSTM (Long Short-Term Memory) & Recurrent Network & \citet{hochreiter1997long, kraft2026engression} \\
EnGRU & GRU (Gated Recurrent Unit) & Recurrent Network & \citet{chung2014empirical} \\
EnBlockRNN & BlockRNN (Block Recurrent Neural Network) & Recurrent Network & \citet{darts2020} \\
EnNLinear & NLinear (Normalized Linear) & Multilayer Perceptron &  \citet{zeng2023transformers} \\
EnDLinear & DLinear (Decomposition Linear) & Multilayer Perceptron & \citet{zeng2023transformers} \\
EnTiDE & TiDE (Time-series Dense Encoder) & Multilayer Perceptron & \citet{das2023longterm} \\
EnTSMixer & TSMixer (Time-Series Mixer) & Multilayer Perceptron & \citet{ekambaram2023tsmixer} \\
EnBEATS & N-BEATS (Neural Basis Expansion Analysis for Time Series) & Multilayer Perceptron & \citet{oreshkin2020nbeats} \\
EnHiTS & N-HiTS (Neural Hierarchical Interpolation for Time Series) & Multilayer Perceptron & \citet{challu2023nhits} \\
EnTCN & TCN (Temporal Convolutional Network) & Convolutional Network & \citet{bai2018empirical} \\
EnTFT & TFT (Temporal Fusion Transformer) & Transformer / Attention-based & \citet{lim2021temporal} \\
EnTransformer & Vanilla Transformer & Transformer / Attention-based & \citet{vaswani2017attention, pathak2026entransformer} \\
\bottomrule
\end{tabular}
\end{adjustbox}
\end{table*}

\subsection{Hyperparameter Configurations}
We conduct the benchmark experiments by utilizing a subset of 11 models from Table~\ref{tab:forecasting_models}. For all experiments, the historical look-back window and prediction horizon are both set to $24$ time steps (hours). The time series are normalized before modeling, and temporal covariates (e.g., hour-of-day, day-of-week, and sequential lags) are extracted and provided to the models to capture seasonal patterns \citep{pathak2026entransformer}. For \texttt{EngressionTS}, we utilize pre-additive noise from a uniform distribution ($\epsilon \sim \mathcal{U}(-\sigma, \sigma)$) injected directly into the normalized inputs with the noise scaling parameter $\sigma$ fixed to $1$. During training, the historical context is replicated $M_{\text{train}}=2$ times to compute the energy score penalty during backpropagation, which is consistent with prior literature \citep{shen2025engression, pathak2026deep}. During inference, we sample $M_{\text{infer}}=100$ independent trajectories from the trained engression models to construct the empirical predictive distribution. 

To guarantee a fair and transparent evaluation, we train and evaluate all models with different probabilistic methods using a consistent set of configurations. All model hyperparameters were set to their default values as specified by \texttt{Darts} Python library \citep{darts2020}. For each model, training was performed utilizing the Adam optimizer with a learning rate of $10^{-3}$, a batch size of $512$ for Solar, and $128$ for Electricity and Traffic, and $30$ training epochs across all datasets. During post-training, we repeat the inference with 10 distinct random seeds, and the final results are reported as the mean and standard deviation of the metrics across these seeds. 

\subsection{Evaluation Metrics}\label{appendix:metrics}
Our evaluation setup comprises a suite of five metrics to assess both probabilistic forecasting performance and point forecasting accuracy. Let $\left\{\widehat{y}_t\right\}_{t=T+1}^{T+q}$ denote the forecasted values and $\left\{y_t\right\}_{t=T+1}^{T+q}$ be the corresponding ground truth, where $q$ is the forecast horizon and $t=\{1,\ldots,T\}$ denotes the time indices for the observed data $\left\{y_t\right\}_{t=1}^T$. The mathematical formulation of the metrics are as follows.

\begin{itemize}
\item \textbf{Mean Absolute Error (MAE)}: MAE is the mean of the absolute differences between the actual observations ($y_t$) and the forecasted values ($\widehat{y}_t$).
$$\text{MAE} = \frac{1}{q} \sum_{t=T+1}^{T+q} \left|\widehat{y}_t - y_t\right|.$$

    \item \textbf{Root Mean Squared Error (RMSE)}: RMSE is the square root of the average of squared differences between observed and predicted values. 
$$\text{RMSE} = \sqrt{\frac{1}{q} \sum_{t=T+1}^{T+q} \left(\widehat{y}_t - y_t\right)^2}.$$
    
    \item \textbf{Continuous Ranked Probability Score (CRPS)}: The CRPS \citep{gneiting2007strictly} generalizes the MAE to the case of probabilistic forecasts. It measures the difference between the predicted cumulative distribution function (CDF) and the empirical CDF of the observation.
    $$\text{CRPS}(F, y) = \frac{1}{q}\sum_{t=T+1}^{T+q} \int_{\mathbb{R}}\left(F_t(y) - \mathbb{I}_{y \geq y_t}\right)^2 \ dy,$$
where $F_t$ denotes the predicted probability distribution at time $t$.

    \item \textbf{Prediction Interval Coverage Probability (PICP)}: PICP is the proportion of times the true values fall within the predicted intervals:
$$\text{PICP} = \frac{1}{q} \sum_{t=T+1}^{T+q} \mathbb{I}_{\left\{\widehat{f}_{t}^{L}\leq y_t\leq \widehat{f}_{t}^{U}\right\}},$$
where $\mathbb{I}$ denotes the indicator function, $y_t$ is the true value, and $\left[\widehat{f}_{t}^{L}, \widehat{f}_{t}^{U}\right]$ is the prediction interval for forecast $t$. For a $95\%$ prediction interval, the best coverage score is $0.95$.




\item \textbf{Mean Prediction Interval Width (MPIW):} MPIW assesses forecast sharpness by computing the average width of the prediction interval. A higher value indicates wide intervals, which are uninformative for decision making. In practice, lower MPIW values with nominal coverage (PICP) is preferred. The metric is defined as:
$$\text{MPIW} = \frac{1}{q}\sum_{t=T+1}^{T+q}\left(\widehat{f}_{t}^{U}-\widehat{f}_{t}^{L}\right).$$

\end{itemize}

\section{Software Structure and Implementation}\label{appendix:software}

\texttt{EngressionTS} is an open-source Python framework built with a PyTorch Lightning backend. Its implementation is organized into four main submodules: \texttt{base}, which contains the core module wrappers; \texttt{losses}, which defines the training objectives; \texttt{noise}, which handles the stochastic noise injection; and \texttt{models}, which contains the different model implementations. Currently, it wraps around the deep learning models from \texttt{Darts} library \citep{darts2020}. In the future, we plan to include support for engression on \texttt{neuralforecast}\footnote{\url{https://github.com/nixtla/neuralforecast}} models.

The core training and inference logic is wrapped within the \texttt{EngressionPLModule}, which builds on Darts' \texttt{PLForecastingModule}. During the forward pass, historical inputs are replicated $M$ times along the batch dimension. Pre-additive noise is sampled from the \texttt{NOISE\_REGISTRY} (e.g., Gaussian or Uniform) and applied to these historical replicas, leaving future and static covariates unperturbed. Users can configure noise parameters such as scale and distribution (\texttt{noise\_std}, \texttt{noise\_type}), training and inference sample sizes, non-negativity constraints (\texttt{clip\_preds}), PyTorch-Lightning trainer settings, optimizers, learning rate schedulers, input and output window lengths, and model-specific hyperparameters. The \texttt{NOISE\_REGISTRY} also supports custom noise modules built with standard PyTorch components.

The in-sample predictions during training are reshaped into a tensor of shape $(M, \text{Batch}, q, N)$ and passed to the empirical energy score loss function. The parameter \texttt{num\_samples\_train} specifies the number of stochastic samples $M$ generated during training, which is typically set to $2$ \citep{shen2025engression, pathak2026deep}. The energy score function computes the distance between each generated forecast and the ground truth, and the pairwise distance between generated forecast samples. These calculations are performed using vectorized PyTorch operations and averaged across the batch and forecast dimensions.

During inference, the hyperparameter \texttt{num\_samples} determines the number of independently generated forecast trajectories ($M_{\text{infer}}$). Rather than collapsing these into a single point forecast, the module returns an empirical predictive distribution as a forecast ensemble, formatted as a Darts \texttt{TimeSeries} object with shape $(q, N, M_{\text{infer}})$. This abstraction ensures that the stochastic inference process works natively across all forecasting models without requiring any model-specific logic or modifications. Point forecasts can be generated by computing, for example, the mean or median of the forecast ensemble along the samples (last) dimension, and specific quantiles can be extracted to produce prediction intervals.

\section{Experimental Results}\label{appendix:results}
\begin{figure*}[!t]
\centering\includegraphics[width=0.9\linewidth]{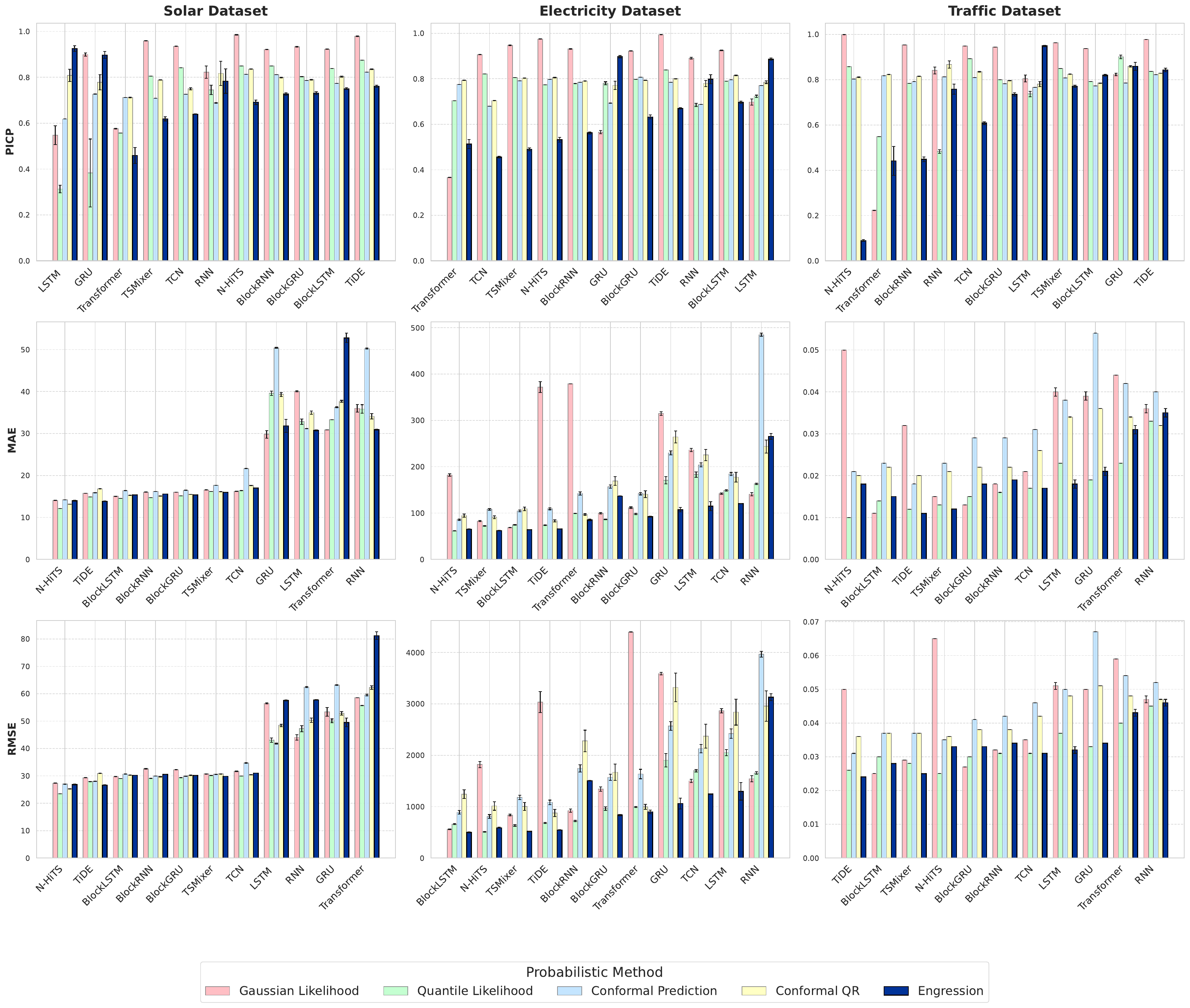}
\caption{{\small PICP (top), MAE (middle), and RMSE (bottom) across models and probabilistic methods for the Solar, Electricity, and Traffic datasets. The models in the X-axis are sorted based on the average scores attained across all 5 paradigms (GL, QR, CP, CQR, and engression).}}\label{fig:picp_mae_rmse_results}
\end{figure*}

\begin{figure*}[!t]
\centering\includegraphics[width=\linewidth]{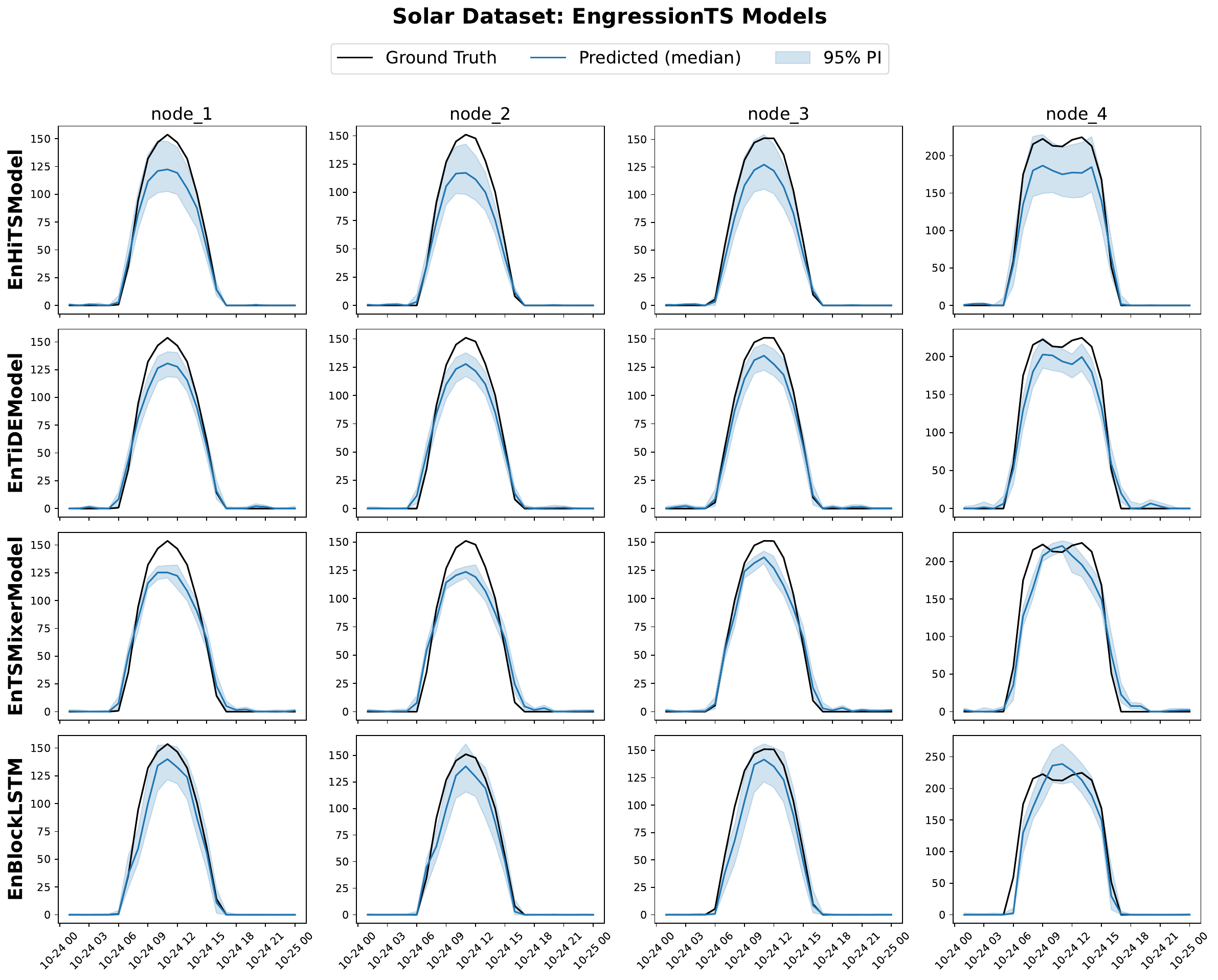}
\caption{Test window 5 of Solar dataset: comparison of predictions against ground truth for four engression-based TS models on first four nodes.}\label{fig:solar_results}
\end{figure*}

\begin{figure*}[!t]
\centering\includegraphics[width=\linewidth]{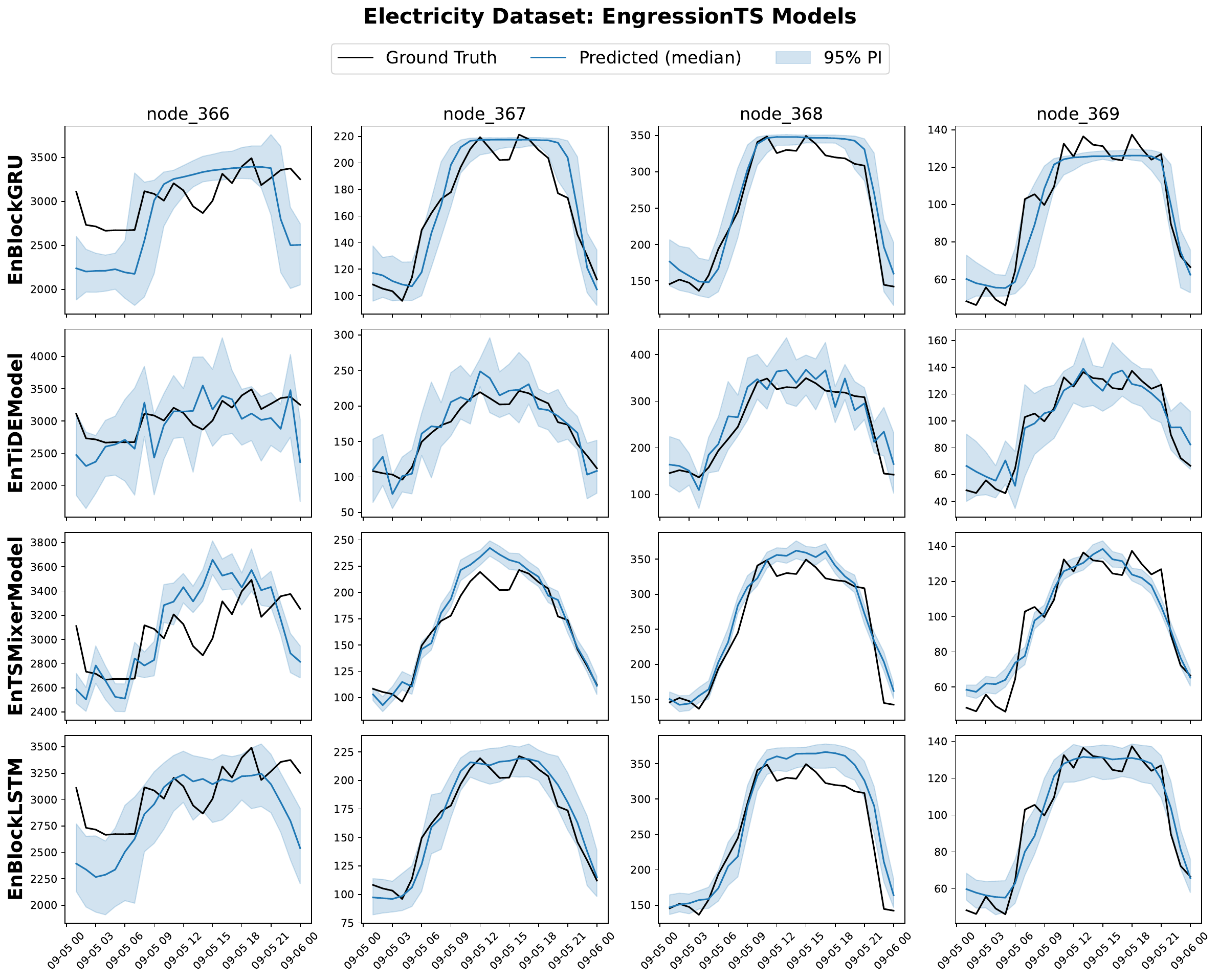}
\caption{Test window 5 of Electricity dataset: comparison of predictions against ground truth for four engression-based TS models on last four nodes.}\label{fig:elec_results}
\end{figure*}

\begin{figure*}[!t]
\centering\includegraphics[width=\linewidth]{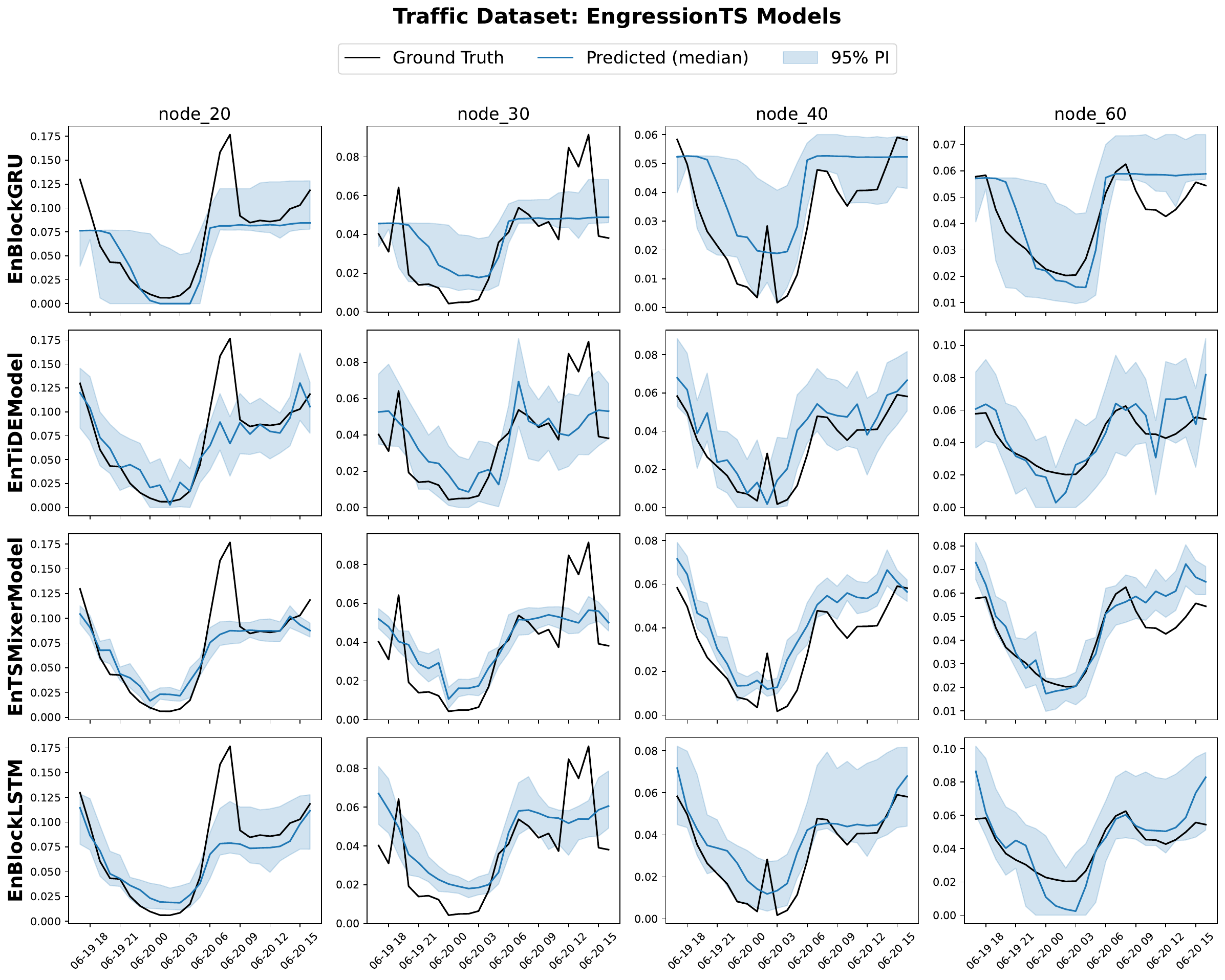}
\caption{Test window 5 of Traffic dataset: comparison of predictions against ground truth for four engression-based TS models on the 20\textsuperscript{th}, 30\textsuperscript{th}, 40\textsuperscript{th}, and 60\textsuperscript{th} nodes.}\label{fig:traffic_results}
\end{figure*}

The quantitative results for the Solar, Electricity, and Traffic datasets obtained across all models and metrics are presented in Tables~\ref{table:solar_results}-\ref{table:traffic_results}. Additionally, visual comparisons for CRPS and MPIW are illustrated in Figure 1 (main text) and the remaining metrics are presented in Figure~\ref{fig:picp_mae_rmse_results}. Lower values indicate better performance for CRPS, MAE, and RMSE, while higher PICP values (approaching the nominal 95\% target) indicate superior empirical coverage. For MPIW, lower values without coverage degradation indicates sharper and more informative prediction intervals. Finally, forecasts and prediction intervals generated by specific engression models on the Solar, Electricity, and Traffic datasets are provided in Figures~\ref{fig:solar_results}, \ref{fig:elec_results}, and \ref{fig:traffic_results}, respectively.

Across all three benchmarks, \texttt{EngressionTS} demonstrates highly competitive probabilistic forecasting performance compared to existing methods, attaining an optimal balance between predictive calibration and sharpness. Engression consistently achieves the lowest (best) CRPS across the vast majority of architectures. For example, the engression-augmented frameworks dominate on the CRPS metric in $\sim 64\%$ and $\sim 73\%$ of the cases for the Solar and Electricity datasets, respectively. Furthermore, engression produces remarkably sharp uncertainty estimates. While other methods struggle with excessive variance, engression achieves the lowest MPIW on a significant portion of the models without catastrophically degrading coverage. The Multiple Comparisons with the Best (MCB) test with respect to CRPS and MPIW metrics (Figure 2; main text) further corroborates this by consistently ranking engression-based models at the top.

Likelihood-based methods (such as Gaussian likelihood) often output overly wide intervals and suffer in CRPS because they enforce strict parametric assumptions on the underlying data distribution, which are frequently violated by real-world time series. Conversely, conformal prediction techniques (CP and CQR) reliably secure high empirical coverage (PICP), but they frequently achieve this at the cost of generating extremely wide, uninformative prediction intervals. Moreover, the calibration step in CP and CQR incurs a substantial computational overhead after the base model is trained. Finally, methods such as quantile regression often suffer from quantile crossing, where the independently estimated quantiles do not respect their monotonicity constraints. Engression overcomes these limitations: by optimizing the energy score over stochastic trajectories, it bypasses restrictive parametric assumptions and avoids the overly conservative bounds and computational overhead of post-hoc conformal calibration.

Although Engression is fundamentally a generative probabilistic framework, it does not sacrifice deterministic accuracy. Point predictions extracted via the empirical median of the sampled trajectories remain highly competitive. Particularly in the Solar and Traffic datasets, Engression frequently achieves the lowest MAE and RMSE on architectures like TiDE and TSMixer, showing that the generative noise-injection mechanism acts as an effective regularizer that can improve point-forecast accuracy over standard deterministic baselines.

\section*{Data and Code Availability Statement}
All the data and code used in this study are made publicly available at \url{https://github.com/anushatomar13/engressionts} for reproducibility. \texttt{EngressionTS} Python package is made available at \url{https://pypi.org/project/engressionts/}.

\begin{table*}[!ht]
    \centering
  \caption{Performance comparison on the Solar dataset across probabilistic forecasting methods and models. Best results are \underline{\textbf{highlighted}} and color-coded based on the metric. The mean and standard deviation for each metric is obtained by using 10 different seeds during inference. Color coding: CRPS (\textcolor{crpscolor}{red}), PICP (\textcolor{picppurplecolor}{purple}), MPIW (\textcolor{mpiwcolor}{orange}), MAE (\textcolor{maecolor}{pink}), and RMSE (\textcolor{rmsecolor}{green}).}
    \begin{adjustbox}{width=\textwidth}
    \begin{tabular}{ccccccccccccc}
    \toprule
        Method & Metrics & BlockRNN & BlockLSTM & BlockGRU & N-HiTS & RNN & LSTM  & GRU & TCN & TiDE & TSMixer & Transformer \\ \midrule
        Gaussian Likelihood & CRPS & $\ms{0.394}{0.000}$ & $\crpsbest{0.370}{0.001}$ & $\ms{0.385}{0.001}$ & $\ms{0.366}{0.001}$ & $\crpsbest{0.828}{0.022}$ & $\ms{1.152}{0.011}$ & $\ms{1.482}{0.097}$ & $\crpsbest{0.416}{0.001}$ & $\ms{0.402}{0.001}$ & $\ms{0.451}{0.000}$ & $\ms{1.010}{0.000}$ \\ 
        ~ & PICP & $\picpbest{0.921}{0.001}$ & $\picpbest{0.923}{0.001}$ & $\picpbest{0.933}{0.001}$ & $\picpbest{0.985}{0.001}$ & $\picpbest{0.822}{0.027}$ & $\ms{0.548}{0.041}$ & $\picpbest{0.899}{0.007}$ & $\picpbest{0.935}{0.001}$ & $\picpbest{0.979}{0.001}$ & $\picpbest{0.959}{0.001}$ & $\ms{0.576}{0.001}$ \\ 
        ~ & MPIW & $\ms{62.402}{0.173}$ & $\ms{56.895}{0.181}$ & $\ms{58.422}{0.158}$ & $\ms{64.048}{0.175}$ & $\ms{115.620}{2.141}$ & $\mpiwbest{49.985}{1.476}$ & $\ms{109.355}{1.908}$ & $\ms{65.145}{0.178}$ & $\ms{69.002}{0.174}$ & $\ms{68.883}{0.195}$ & $\mpiwbest{2.631}{0.007}$ \\ 
        ~ & MAE & $\ms{16.031}{0.032}$ & $\ms{15.033}{0.029}$ & $\ms{16.012}{0.028}$ & $\ms{14.070}{0.016}$ & $\ms{36.001}{0.880}$ & $\ms{40.039}{0.123}$ & $\maebest{29.817}{0.909}$ & $\maebest{16.196}{0.029}$ & $\ms{15.747}{0.024}$ & $\ms{16.551}{0.026}$ & $\maebest{30.837}{0.002}$ \\ 
        ~ & RMSE & $\ms{32.628}{0.131}$ & $\ms{29.755}{0.071}$ & $\ms{32.278}{0.102}$ & $\ms{27.308}{0.060}$ & $\rmsebest{44.028}{0.995}$ & $\ms{56.504}{0.155}$ & $\ms{53.409}{1.580}$ & $\ms{31.656}{0.100}$ & $\ms{29.300}{0.068}$ & $\ms{30.710}{0.084}$ & $\ms{58.516}{0.003}$ \\ \midrule
        Quantile Likelihood & CRPS & $\ms{0.388}{0.001}$ & $\ms{0.382}{0.000}$ & $\ms{0.407}{0.001}$ & $\ms{0.301}{0.000}$ & $\ms{0.852}{0.029}$ & $\ms{0.848}{0.022}$ & $\ms{1.053}{0.024}$ & $\ms{0.483}{0.001}$ & $\ms{0.386}{0.001}$ & $\ms{0.438}{0.000}$ & $\crpsbest{0.998}{0.001}$ \\ 
        ~ & PICP & $\ms{0.849}{0.000}$ & $\ms{0.839}{0.000}$ & $\ms{0.803}{0.001}$ & $\ms{0.849}{0.000}$ & $\ms{0.745}{0.020}$ & $\ms{0.313}{0.016}$ & $\ms{0.383}{0.148}$ & $\ms{0.842}{0.000}$ & $\ms{0.875}{0.000}$ & $\ms{0.805}{0.000}$ & $\ms{0.557}{0.000}$ \\ 
        ~ & MPIW & $\ms{37.772}{0.007}$ & $\ms{35.863}{0.008}$ & $\ms{35.781}{0.003}$ & $\ms{30.991}{0.002}$ & $\ms{88.107}{2.222}$ & $\ms{63.717}{1.253}$ & $\mpiwbest{67.940}{2.282}$ & $\ms{53.949}{0.013}$ & $\ms{41.692}{0.005}$ & $\ms{39.417}{0.003}$ & $\ms{33.166}{0.004}$ \\ 
        ~ & MAE & $\maebest{14.692}{0.014}$ & $\maebest{14.542}{0.014}$ & $\maebest{15.127}{0.013}$ & $\maebest{12.147}{0.012}$ & $\ms{35.828}{1.034}$ & $\ms{32.824}{0.655}$ & $\ms{39.607}{0.515}$ & $\ms{16.408}{0.018}$ & $\ms{14.863}{0.016}$ & $\ms{16.193}{0.015}$ & $\ms{33.268}{0.028}$ \\ 
        ~ & RMSE & $\rmsebest{29.093}{0.037}$ & $\rmsebest{29.032}{0.041}$ & $\rmsebest{29.332}{0.038}$ & $\rmsebest{23.443}{0.032}$ & $\ms{47.268}{1.139}$ & $\ms{43.067}{0.796}$ & $\ms{50.107}{0.689}$ & $\rmsebest{29.910}{0.017}$ & $\ms{27.925}{0.029}$ & $\ms{30.160}{0.036}$ & $\rmsebest{55.665}{0.071}$ \\ \midrule
        Conformal Prediction & CRPS & $\ms{0.406}{0.000}$ & $\ms{0.450}{0.000}$ & $\ms{0.428}{0.000}$ & $\ms{0.386}{0.001}$ & $\ms{1.496}{0.002}$ & $\ms{0.904}{0.001}$ & $\ms{1.516}{0.002}$ & $\ms{0.506}{0.001}$ & $\ms{0.412}{0.000}$ & $\ms{0.495}{0.000}$ & $\ms{1.109}{0.002}$ \\ 
        ~ & PICP & $\ms{0.811}{0.000}$ & $\ms{0.774}{0.000}$ & $\ms{0.786}{0.000}$ & $\ms{0.813}{0.000}$ & $\ms{0.688}{0.001}$ & $\ms{0.618}{0.000}$ & $\ms{0.727}{0.001}$ & $\ms{0.726}{0.000}$ & $\ms{0.823}{0.000}$ & $\ms{0.708}{0.000}$ & $\ms{0.712}{0.000}$ \\ 
        ~ & MPIW & $\ms{37.926}{0.002}$ & $\ms{36.417}{0.002}$ & $\ms{37.343}{0.002}$ & $\ms{34.985}{0.002}$ & $\ms{111.414}{0.003}$ & $\ms{71.103}{0.002}$ & $\ms{110.193}{0.003}$ & $\ms{48.516}{0.002}$ & $\ms{41.167}{0.002}$ & $\ms{36.518}{0.002}$ & $\ms{73.929}{0.003}$ \\ 
        ~ & MAE & $\ms{16.175}{0.015}$ & $\ms{16.399}{0.010}$ & $\ms{16.486}{0.019}$ & $\ms{14.201}{0.014}$ & $\ms{50.250}{0.123}$ & $\ms{31.163}{0.072}$ & $\ms{50.392}{0.118}$ & $\ms{21.628}{0.039}$ & $\ms{15.903}{0.025}$ & $\ms{17.638}{0.012}$ & $\ms{36.222}{0.123}$ \\ 
        ~ & RMSE & $\ms{29.928}{0.039}$ & $\ms{30.685}{0.046}$ & $\ms{29.927}{0.043}$ & $\ms{27.019}{0.046}$ & $\ms{62.444}{0.192}$ & $\rmsebest{41.793}{0.139}$ & $\ms{63.183}{0.164}$ & $\ms{34.672}{0.117}$ & $\ms{27.983}{0.050}$ & $\ms{30.575}{0.038}$ & $\ms{59.523}{0.298}$ \\ \midrule
        Conformalized QR & CRPS & $\ms{0.401}{0.001}$ & $\ms{0.409}{0.000}$ & $\ms{0.420}{0.000}$ & $\ms{0.323}{0.001}$ & $\ms{1.009}{0.022}$ & $\ms{1.022}{0.013}$ & $\ms{1.167}{0.014}$ & $\ms{0.544}{0.001}$ & $\ms{0.456}{0.001}$ & $\ms{0.436}{0.001}$ & $\ms{1.156}{0.002}$ \\ 
        ~ & PICP & $\ms{0.799}{0.002}$ & $\ms{0.803}{0.002}$ & $\ms{0.789}{0.002}$ & $\ms{0.836}{0.001}$ & $\ms{0.817}{0.053}$ & $\ms{0.808}{0.027}$ & $\ms{0.778}{0.033}$ & $\ms{0.750}{0.005}$ & $\ms{0.835}{0.001}$ & $\ms{0.788}{0.001}$ & $\picpbest{0.712}{0.001}$ \\ 
        ~ & MPIW & $\ms{34.523}{0.081}$ & $\ms{33.886}{0.106}$ & $\ms{35.001}{0.077}$ & $\ms{35.804}{0.038}$ & $\ms{77.556}{0.853}$ & $\ms{85.062}{0.829}$ & $\ms{91.614}{0.685}$ & $\ms{56.293}{0.073}$ & $\ms{45.071}{0.073}$ & $\ms{35.475}{0.075}$ & $\ms{71.700}{0.047}$ \\ 
        ~ & MAE & $\ms{15.087}{0.019}$ & $\ms{15.259}{0.027}$ & $\ms{15.491}{0.023}$ & $\ms{13.148}{0.019}$ & $\ms{34.093}{0.660}$ & $\ms{34.961}{0.373}$ & $\ms{39.298}{0.478}$ & $\ms{17.624}{0.047}$ & $\ms{16.842}{0.031}$ & $\ms{16.137}{0.028}$ & $\ms{37.644}{0.253}$ \\ 
        ~ & RMSE & $\ms{29.683}{0.062}$ & $\ms{30.250}{0.090}$ & $\ms{30.217}{0.075}$ & $\ms{25.244}{0.059}$ & $\ms{50.362}{0.766}$ & $\ms{48.460}{0.461}$ & $\ms{52.861}{0.612}$ & $\ms{30.375}{0.069}$ & $\ms{30.909}{0.084}$ & $\ms{30.703}{0.098}$ & $\ms{62.257}{0.636}$ \\ \midrule
        Engression & CRPS & $\crpsbest{0.377}{0.002}$ & $\ms{0.375}{0.002}$ & $\crpsbest{0.367}{0.002}$ & $\crpsbest{0.290}{0.002}$ & $\ms{0.926}{0.015}$ & $\crpsbest{0.792}{0.028}$ & $\crpsbest{0.777}{0.034}$ & $\ms{0.437}{0.002}$ & $\crpsbest{0.329}{0.004}$ & $\crpsbest{0.413}{0.002}$ & $\ms{1.451}{0.035}$ \\ 
        ~ & PICP & $\ms{0.727}{0.006}$ & $\ms{0.750}{0.004}$ & $\ms{0.731}{0.006}$ & $\ms{0.691}{0.010}$ & $\ms{0.783}{0.053}$ & $\picpbest{0.925}{0.013}$ & $\ms{0.897}{0.015}$ & $\ms{0.639}{0.002}$ & $\ms{0.761}{0.005}$ & $\ms{0.619}{0.009}$ & $\ms{0.459}{0.035}$ \\ 
        ~ & MPIW & $\mpiwbest{26.147}{0.481}$ & $\mpiwbest{22.432}{0.401}$ & $\mpiwbest{22.366}{0.652}$ & $\mpiwbest{26.168}{0.969}$ & $\mpiwbest{58.530}{19.894}$ & $\ms{110.588}{6.790}$ & $\ms{112.990}{3.275}$ & $\mpiwbest{17.295}{0.236}$ & $\mpiwbest{23.018}{0.761}$ & $\mpiwbest{13.821}{0.384}$ & $\ms{62.281}{6.411}$ \\ 
        ~ & MAE & $\ms{15.582}{0.018}$ & $\ms{15.366}{0.012}$ & $\ms{15.351}{0.031}$ & $\ms{13.982}{0.128}$ & $\maebest{30.911}{0.114}$ & $\maebest{30.766}{0.129}$ & $\ms{31.800}{1.542}$ & $\ms{17.022}{0.050}$ & $\maebest{13.858}{0.064}$ & $\maebest{15.946}{0.042}$ & $\ms{52.763}{1.123}$ \\ 
        ~ & RMSE & $\ms{30.512}{0.027}$ & $\ms{30.241}{0.026}$ & $\ms{30.128}{0.048}$ & $\ms{26.811}{0.218}$ & $\ms{57.745}{0.146}$ & $\ms{57.599}{0.181}$ & $\rmsebest{49.506}{1.564}$ & $\ms{31.029}{0.019}$ & $\rmsebest{26.616}{0.156}$ & $\rmsebest{29.830}{0.032}$ & $\ms{81.206}{1.402}$ \\ \bottomrule
    \end{tabular}
    \end{adjustbox}
    \label{table:solar_results}
\end{table*}

\begin{table*}[!ht]
    \centering
  \caption{Performance comparison on the Electricity dataset across probabilistic forecasting methods and models. Best results are \underline{\textbf{highlighted}} and color-coded based on the metric. The mean and standard deviation for each metric is obtained by using 10 different seeds during inference. Color coding: CRPS (\textcolor{crpscolor}{red}), PICP (\textcolor{picppurplecolor}{purple}), MPIW (\textcolor{mpiwcolor}{orange}), MAE (\textcolor{maecolor}{pink}), and RMSE (\textcolor{rmsecolor}{green}).}
    \begin{adjustbox}{width=\textwidth}
    \begin{tabular}{ccccccccccccc}
    \toprule
        Method & Metrics & BlockRNN & BlockLSTM & BlockGRU & N-HiTS & RNN & LSTM  & GRU & TCN & TiDE & TSMixer & Transformer \\ \midrule
        Gaussian Likelihood & CRPS & $\ms{0.051}{0.001}$ & $\ms{0.025}{0.000}$ & $\ms{0.071}{0.002}$ & $\ms{0.163}{0.002}$ & $\crpsbest{0.111}{0.004}$ & $\ms{0.226}{0.007}$ & $\ms{0.348}{0.005}$ & $\ms{0.084}{0.001}$ & $\ms{0.222}{0.008}$ & $\ms{0.045}{0.001}$ & $\ms{0.473}{0.000}$ \\ 
        ~ & PICP & $\picpbest{0.931}{0.001}$ & $\picpbest{0.925}{0.001}$ & $\picpbest{0.922}{0.001}$ & $\picpbest{0.975}{0.001}$ & $\picpbest{0.890}{0.004}$ & $\ms{0.698}{0.014}$ & $\ms{0.565}{0.007}$ & $\picpbest{0.906}{0.001}$ & $\picpbest{0.994}{0.000}$ & $\picpbest{0.947}{0.001}$ & $\ms{0.366}{0.001}$ \\ 
        ~ & MPIW & $\ms{404.783}{2.531}$ & $\ms{306.410}{1.666}$ & $\ms{409.722}{3.114}$ & $\ms{814.904}{5.574}$ & $\ms{539.409}{21.805}$ & $\mpiwbest{405.823}{11.822}$ & $\mpiwbest{321.634}{6.104}$ & $\ms{501.231}{3.190}$ & $\ms{2579.873}{17.570}$ & $\ms{395.189}{2.836}$ & $\mpiwbest{157.843}{0.605}$ \\ 
        ~ & MAE & $\ms{99.519}{1.405}$ & $\ms{68.524}{0.221}$ & $\ms{111.495}{1.823}$ & $\ms{182.192}{2.743}$ & $\maebest{140.780}{3.480}$ & $\ms{236.150}{3.510}$ & $\ms{314.809}{3.832}$ & $\ms{142.195}{1.548}$ & $\ms{371.580}{11.715}$ & $\ms{82.812}{0.882}$ & $\ms{378.863}{0.290}$ \\ 
        ~ & RMSE & $\ms{923.014}{30.794}$ & $\ms{560.064}{5.722}$ & $\ms{1343.184}{41.152}$ & $\ms{1819.223}{58.394}$ & $\rmsebest{1543.091}{57.735}$ & $\ms{2864.242}{42.452}$ & $\ms{3589.523}{24.103}$ & $\ms{1499.835}{31.973}$ & $\ms{3034.780}{205.646}$ & $\ms{834.410}{17.593}$ & $\ms{4397.811}{3.858}$ \\ \midrule
        Quantile Likelihood & CRPS & $\crpsbest{0.041}{0.000}$ & $\ms{0.029}{0.000}$ & $\ms{0.053}{0.001}$ & $\crpsbest{0.035}{0.000}$ & $\ms{0.148}{0.001}$ & $\ms{0.172}{0.005}$ & $\ms{0.142}{0.008}$ & $\ms{0.086}{0.002}$ & $\ms{0.043}{0.001}$ & $\ms{0.030}{0.001}$ & $\ms{0.074}{0.001}$ \\ 
        ~ & PICP & $\ms{0.779}{0.000}$ & $\ms{0.790}{0.000}$ & $\ms{0.797}{0.000}$ & $\ms{0.773}{0.000}$ & $\ms{0.685}{0.007}$ & $\ms{0.723}{0.005}$ & $\ms{0.780}{0.007}$ & $\ms{0.821}{0.000}$ & $\ms{0.839}{0.000}$ & $\ms{0.805}{0.000}$ & $\ms{0.704}{0.000}$ \\ 
        ~ & MPIW & $\ms{274.345}{0.089}$ & $\ms{250.568}{0.047}$ & $\ms{336.669}{0.095}$ & $\ms{214.100}{0.134}$ & $\mpiwbest{362.331}{6.746}$ & $\ms{423.781}{11.844}$ & $\ms{456.241}{10.670}$ & $\ms{456.914}{0.262}$ & $\ms{256.206}{0.098}$ & $\ms{245.930}{0.027}$ & $\ms{230.289}{0.028}$ \\ 
        ~ & MAE & $\maebest{86.261}{0.633}$ & $\ms{74.836}{0.641}$ & $\ms{98.055}{1.645}$ & $\maebest{61.479}{0.419}$ & $\ms{163.051}{1.504}$ & $\ms{183.164}{5.827}$ & $\ms{170.660}{7.720}$ & $\ms{148.786}{1.355}$ & $\ms{73.821}{0.732}$ & $\ms{72.312}{0.808}$ & $\ms{99.266}{0.706}$ \\ 
        ~ & RMSE & $\rmsebest{718.926}{14.645}$ & $\ms{659.616}{10.360}$ & $\ms{960.419}{32.935}$ & $\rmsebest{510.194}{8.458}$ & $\ms{1654.492}{22.706}$ & $\ms{2049.728}{61.205}$ & $\ms{1903.516}{128.192}$ & $\ms{1699.276}{23.897}$ & $\ms{681.564}{12.153}$ & $\ms{631.618}{17.205}$ & $\ms{992.861}{12.870}$ \\ \midrule
        Conformal Prediction & CRPS & $\ms{0.105}{0.002}$ & $\ms{0.043}{0.001}$ & $\ms{0.087}{0.002}$ & $\ms{0.041}{0.001}$ & $\ms{0.786}{0.005}$ & $\ms{0.174}{0.003}$ & $\ms{0.204}{0.003}$ & $-$ & $\ms{0.060}{0.001}$ & $\ms{0.061}{0.001}$ & $\ms{0.110}{0.002}$ \\ 
        ~ & PICP & $\ms{0.785}{0.000}$ & $\ms{0.795}{0.000}$ & $\ms{0.807}{0.000}$ & $\ms{0.797}{0.000}$ & $\ms{0.687}{0.000}$ & $\ms{0.770}{0.000}$ & $\ms{0.693}{0.000}$ & $\ms{0.679}{0.000}$ & $\ms{0.785}{0.000}$ & $\ms{0.791}{0.000}$ & $\ms{0.775}{0.000}$ \\ 
        ~ & MPIW & $\ms{415.634}{0.123}$ & $\ms{307.690}{0.071}$ & $\ms{380.992}{0.105}$ & $\ms{263.376}{0.049}$ & $\ms{992.873}{0.028}$ & $\ms{509.624}{0.069}$ & $\ms{522.830}{0.076}$ & $\ms{471.074}{0.162}$ & $\ms{302.411}{0.080}$ & $\ms{303.327}{0.099}$ & $\ms{365.868}{0.151}$ \\ 
        ~ & MAE & $\ms{156.830}{3.489}$ & $\ms{104.784}{1.797}$ & $\ms{141.680}{2.539}$ & $\ms{85.466}{1.375}$ & $\ms{484.511}{3.552}$ & $\ms{203.824}{4.200}$ & $\ms{229.799}{4.182}$ & $\ms{184.123}{3.501}$ & $\ms{109.113}{2.074}$ & $\ms{108.084}{1.866}$ & $\ms{142.343}{3.502}$ \\ 
        ~ & RMSE & $\ms{1746.793}{70.311}$ & $\ms{887.641}{35.556}$ & $\ms{1569.931}{59.561}$ & $\ms{807.879}{36.671}$ & $\ms{3965.689}{54.119}$ & $\ms{2421.241}{90.946}$ & $\ms{2569.424}{86.247}$ & $\ms{2129.164}{82.073}$ & $\ms{1083.488}{43.330}$ & $\ms{1180.948}{42.175}$ & $\ms{1634.577}{91.686}$ \\ \midrule
        Conformalized QR & CRPS & $\ms{0.096}{0.003}$ & $\ms{0.052}{0.001}$ & $\ms{0.081}{0.002}$ & $\ms{0.046}{0.001}$ & $\ms{0.193}{0.006}$ & $\ms{0.183}{0.005}$ & $\ms{0.223}{0.005}$ & $-$ & $\ms{0.038}{0.001}$ & $\ms{0.044}{0.001}$ & $\ms{0.057}{0.001}$ \\ 
        ~ & PICP & $\ms{0.790}{0.001}$ & $\ms{0.815}{0.001}$ & $\ms{0.793}{0.001}$ & $\ms{0.805}{0.001}$ & $\ms{0.779}{0.014}$ & $\ms{0.785}{0.006}$ & $\ms{0.771}{0.018}$ & $\ms{0.704}{0.001}$ & $\ms{0.800}{0.001}$ & $\ms{0.803}{0.001}$ & $\picpbest{0.793}{0.001}$ \\ 
        ~ & MPIW & $\ms{417.330}{0.982}$ & $\ms{320.654}{0.814}$ & $\ms{404.189}{1.789}$ & $\ms{285.753}{1.314}$ & $\ms{563.955}{9.348}$ & $\ms{528.366}{2.735}$ & $\ms{562.875}{4.433}$ & $\ms{499.071}{2.249}$ & $\ms{266.956}{1.038}$ & $\ms{290.685}{0.943}$ & $\ms{321.026}{0.890}$ \\ 
        ~ & MAE & $\ms{169.284}{9.359}$ & $\ms{109.328}{3.892}$ & $\ms{140.687}{7.222}$ & $\ms{94.542}{3.656}$ & $\ms{243.861}{14.082}$ & $\ms{225.363}{12.043}$ & $\ms{264.221}{12.778}$ & $\ms{177.455}{10.414}$ & $\ms{83.493}{2.330}$ & $\ms{91.183}{3.129}$ & $\ms{96.563}{1.891}$ \\ 
        ~ & RMSE & $\ms{2278.136}{209.988}$ & $\ms{1244.396}{84.510}$ & $\ms{1668.994}{156.818}$ & $\ms{1013.058}{83.779}$ & $\ms{2957.381}{299.024}$ & $\ms{2840.210}{253.132}$ & $\ms{3320.647}{280.638}$ & $\ms{2369.472}{232.624}$ & $\ms{875.076}{71.341}$ & $\ms{1002.512}{79.177}$ & $\ms{994.446}{49.404}$ \\ \midrule
        Engression & CRPS & $\ms{0.098}{0.001}$ & $\crpsbest{0.023}{0.000}$ & $\crpsbest{0.043}{0.001}$ & $\ms{0.037}{0.001}$ & $\ms{0.250}{0.010}$ & $\crpsbest{0.083}{0.007}$ & $\crpsbest{0.077}{0.003}$ & $\crpsbest{0.075}{0.001}$ & $\crpsbest{0.028}{0.001}$ & $\crpsbest{0.025}{0.001}$ & $\crpsbest{0.056}{0.002}$ \\ 
        ~ & PICP & $\ms{0.562}{0.005}$ & $\ms{0.697}{0.004}$ & $\ms{0.632}{0.009}$ & $\ms{0.532}{0.011}$ & $\ms{0.798}{0.020}$ & $\picpbest{0.886}{0.005}$ & $\picpbest{0.897}{0.006}$ & $\ms{0.456}{0.004}$ & $\ms{0.670}{0.003}$ & $\ms{0.490}{0.006}$ & $\ms{0.513}{0.020}$ \\ 
        ~ & MPIW & $\mpiwbest{187.622}{2.745}$ & $\ms{198.002}{2.660}$ & $\mpiwbest{193.069}{3.168}$ & $\mpiwbest{120.537}{3.759}$ & $\ms{631.533}{18.751}$ & $\ms{545.014}{9.119}$ & $\ms{617.123}{16.284}$ & $\mpiwbest{148.828}{1.599}$ & $\mpiwbest{161.986}{4.109}$ & $\mpiwbest{110.729}{2.716}$ & $\ms{183.700}{15.340}$ \\ 
        ~ & MAE & $\ms{136.479}{0.337}$ & $\maebest{63.929}{0.474}$ & $\maebest{92.189}{0.993}$ & $\ms{65.017}{0.706}$ & $\ms{265.551}{6.501}$ & $\maebest{114.897}{9.944}$ & $\maebest{107.222}{4.748}$ & $\maebest{120.427}{0.420}$ & $\maebest{65.893}{0.424}$ & $\maebest{62.130}{0.431}$ & $\maebest{85.057}{1.683}$ \\ 
        ~ & RMSE & $\ms{1503.009}{7.350}$ & $\rmsebest{498.594}{12.289}$ & $\rmsebest{836.307}{14.400}$ & $\ms{588.074}{9.391}$ & $\ms{3131.088}{66.758}$ & $\rmsebest{1296.192}{169.798}$ & $\rmsebest{1058.551}{105.439}$ & $\rmsebest{1244.933}{7.730}$ & $\rmsebest{544.386}{6.066}$ & $\rmsebest{518.984}{6.066}$ & $\rmsebest{898.475}{34.868}$ \\ \bottomrule
    \end{tabular}
    \end{adjustbox}
\end{table*}

\begin{table*}[!ht]
    \centering
    \caption{Performance comparison on the Traffic dataset across probabilistic forecasting methods and models. Best results are \underline{\textbf{highlighted}} and color-coded based on the metric. The mean and standard deviation for each metric is obtained by using 10 different seeds during inference. Color coding: CRPS (\textcolor{crpscolor}{red}), PICP (\textcolor{picppurplecolor}{purple}), MPIW  (\textcolor{miscolor}{blue}), MAE (\textcolor{maecolor}{pink}), and RMSE (\textcolor{rmsecolor}{green}).}
    \begin{adjustbox}{width=\textwidth}
    \begin{tabular}{ccccccccccccc}
    \toprule
        Method & Metrics & BlockRNN & BlockLSTM & BlockGRU & N-HiTS & RNN & LSTM & GRU & TCN & TiDE & TSMixer & Transformer \\ \midrule

        Gaussian Likelihood & CRPS &
        $\ms{0.158}{0.000}$ &
        $\crpsbest{0.086}{0.000}$ &
        $\crpsbest{0.125}{0.000}$ &
        $\ms{1.673}{0.003}$ &
        $\ms{0.494}{0.012}$ &
        $\ms{0.571}{0.020}$ &
        $\ms{0.546}{0.010}$ &
        $\ms{0.181}{0.000}$ &
        $\ms{0.228}{0.001}$ &
        $\ms{0.118}{0.000}$ &
        $\ms{0.828}{0.000}$ \\

        ~ & PICP &
        $\picpbest{0.953}{0.000}$ &
        $\picpbest{0.937}{0.000}$ &
        $\picpbest{0.944}{0.000}$ &
        $\picpbest{0.999}{0.000}$ &
        $\ms{0.841}{0.015}$ &
        $\ms{0.805}{0.015}$ &
        $\ms{0.823}{0.005}$ &
        $\picpbest{0.949}{0.000}$ &
        $\picpbest{0.978}{0.000}$ &
        $\picpbest{0.963}{0.000}$ &
        $\ms{0.223}{0.001}$ \\

        ~ & MPIW &
        $\ms{0.091}{0.000}$ &
        $\ms{0.057}{0.000}$ &
        $\ms{0.062}{0.000}$ &
        $\ms{0.443}{0.000}$ &
        $\ms{0.136}{0.001}$ &
        $\ms{0.141}{0.001}$ &
        $\ms{0.140}{0.001}$ &
        $\ms{0.094}{0.000}$ &
        $\ms{0.203}{0.000}$ &
        $\ms{0.082}{0.000}$ &
        $\mpiwbest{0.023}{0.000}$ \\


        ~ & MAE &
        $\ms{0.018}{0.000}$ &
        $\maebest{0.011}{0.000}$ &
        $\maebest{0.013}{0.000}$ &
        $\ms{0.050}{0.000}$ &
        $\ms{0.036}{0.001}$ &
        $\ms{0.040}{0.001}$ &
        $\ms{0.039}{0.001}$ &
        $\ms{0.021}{0.000}$ &
        $\ms{0.032}{0.000}$ &
        $\ms{0.015}{0.000}$ &
        $\ms{0.044}{0.000}$ \\

        ~ & RMSE &
        $\ms{0.032}{0.000}$ &
        $\rmsebest{0.025}{0.000}$ &
        $\rmsebest{0.027}{0.000}$ &
        $\ms{0.065}{0.000}$ &
        $\ms{0.047}{0.001}$ &
        $\ms{0.051}{0.001}$ &
        $\ms{0.050}{0.001}$ &
        $\ms{0.035}{0.000}$ &
        $\ms{0.050}{0.000}$ &
        $\ms{0.029}{0.000}$ &
        $\ms{0.059}{0.000}$ \\ \midrule

        Quantile Likelihood & CRPS &
        $\crpsbest{0.153}{0.000}$ &
        $\ms{0.114}{0.000}$ &
        $\ms{0.129}{0.000}$ &
        $\crpsbest{0.126}{0.000}$ &
        $\ms{0.468}{0.005}$ &
        $\ms{0.306}{0.007}$ &
        $\crpsbest{0.206}{0.006}$ &
        $\ms{0.167}{0.000}$ &
        $\ms{0.107}{0.000}$ &
        $\ms{0.094}{0.000}$ &
        $\crpsbest{0.355}{0.000}$ \\

        ~ & PICP &
        $\ms{0.784}{0.000}$ &
        $\ms{0.792}{0.000}$ &
        $\ms{0.800}{0.000}$ &
        $\ms{0.858}{0.000}$ &
        $\ms{0.483}{0.008}$ &
        $\ms{0.737}{0.012}$ &
        $\picpbest{0.901}{0.008}$ &
        $\ms{0.893}{0.000}$ &
        $\ms{0.837}{0.000}$ &
        $\ms{0.845}{0.000}$ &
        $\ms{0.548}{0.000}$ \\

        ~ & MPIW &
        $\ms{0.048}{0.000}$ &
        $\ms{0.043}{0.000}$ &
        $\ms{0.047}{0.000}$ &
        $\ms{0.034}{0.000}$ &
        $\mpiwbest{0.071}{0.001}$ &
        $\mpiwbest{0.067}{0.002}$ &
        $\ms{0.083}{0.001}$ &
        $\ms{0.085}{0.000}$ &
        $\ms{0.043}{0.000}$ &
        $\ms{0.045}{0.000}$ &
        $\mpiwbest{0.034}{0.000}$ \\


        ~ & MAE &
        $\ms{0.016}{0.000}$ &
        $\ms{0.014}{0.000}$ &
        $\ms{0.015}{0.000}$ &
        $\ms{0.030}{0.000}$ &
        $\ms{0.033}{0.001}$ &
        $\ms{0.033}{0.001}$ &
        $\maebest{0.019}{0.000}$ &
        $\ms{0.034}{0.000}$ &
        $\maebest{0.012}{0.000}$ &
        $\ms{0.023}{0.000}$ &
        $\ms{0.023}{0.000}$ \\

        ~ & RMSE &
        $\ms{0.031}{0.000}$ &
        $\ms{0.030}{0.000}$ &
        $\ms{0.030}{0.000}$ &
        $\rmsebest{0.025}{0.000}$ &
        $\ms{0.045}{0.000}$ &
        $\ms{0.037}{0.000}$ &
        $\ms{0.033}{0.000}$ &
        $\ms{0.031}{0.000}$ &
        $\ms{0.026}{0.000}$ &
        $\ms{0.037}{0.000}$ &
        $\ms{0.040}{0.000}$ \\ \midrule

        Conformal Prediction & CRPS &
        $\ms{0.323}{0.000}$ &
        $\ms{0.195}{0.000}$ &
        $\ms{0.360}{0.000}$ &
        $\crpsbest{0.126}{0.000}$ &
        $\ms{0.560}{0.000}$ &
        $\ms{0.470}{0.000}$ &
        $\ms{0.368}{0.000}$ &
        $\ms{0.254}{0.000}$ &
        $\ms{0.231}{0.000}$ &
        $\ms{0.150}{0.000}$ &
        $\ms{0.682}{0.000}$ \\

        ~ & PICP &
        $\ms{0.791}{0.000}$ &
        $\ms{0.773}{0.000}$ &
        $\ms{0.782}{0.000}$ &
        $\ms{0.803}{0.000}$ &
        $\ms{0.812}{0.000}$ &
        $\ms{0.766}{0.000}$ &
        $\ms{0.786}{0.000}$ &
        $\ms{0.809}{0.000}$ &
        $\ms{0.822}{0.000}$ &
        $\ms{0.808}{0.000}$ &
        $\ms{0.818}{0.000}$ \\

        ~ & MPIW &
        $\ms{0.080}{0.000}$ &
        $\ms{0.063}{0.000}$ &
        $\ms{0.078}{0.000}$ &
        $\ms{0.057}{0.000}$ &
        $\ms{0.100}{0.001}$ &
        $\ms{0.116}{0.001}$ &
        $\ms{0.134}{0.001}$ &
        $\ms{0.078}{0.000}$ &
        $\ms{0.053}{0.000}$ &
        $\ms{0.063}{0.000}$ &
        $\ms{0.099}{0.000}$ \\


        ~ & MAE &
        $\ms{0.029}{0.000}$ &
        $\ms{0.023}{0.000}$ &
        $\ms{0.029}{0.000}$ &
        $\maebest{0.021}{0.000}$ &
        $\ms{0.040}{0.000}$ &
        $\ms{0.038}{0.000}$ &
        $\ms{0.054}{0.000}$ &
        $\ms{0.031}{0.000}$ &
        $\maebest{0.018}{0.000}$ &
        $\ms{0.023}{0.000}$ &
        $\ms{0.042}{0.000}$ \\

        ~ & RMSE &
        $\ms{0.042}{0.000}$ &
        $\ms{0.037}{0.000}$ &
        $\ms{0.041}{0.000}$ &
        $\ms{0.035}{0.000}$ &
        $\ms{0.052}{0.000}$ &
        $\ms{0.050}{0.000}$ &
        $\ms{0.067}{0.000}$ &
        $\ms{0.046}{0.000}$ &
        $\rmsebest{0.031}{0.000}$ &
        $\ms{0.037}{0.000}$ &
        $\ms{0.054}{0.000}$ \\ \midrule

        Conformalized QR & CRPS &
        $\ms{0.215}{0.000}$ &
        $\ms{0.204}{0.000}$ &
        $\ms{0.215}{0.000}$ &
        $\ms{0.212}{0.000}$ &
        $\ms{0.423}{0.012}$ &
        $\ms{0.501}{0.004}$ &
        $\ms{0.569}{0.005}$ &
        $\ms{0.386}{0.001}$ &
        $\crpsbest{0.161}{0.000}$ &
        $\crpsbest{0.162}{0.000}$ &
        $\ms{0.573}{0.000}$ \\

        ~ & PICP &
        $\ms{0.815}{0.001}$ &
        $\ms{0.785}{0.001}$ &
        $\ms{0.796}{0.001}$ &
        $\ms{0.811}{0.001}$ &
        $\picpbest{0.867}{0.016}$ &
        $\ms{0.760}{0.011}$ &
        $\ms{0.859}{0.003}$ &
        $\ms{0.835}{0.001}$ &
        $\ms{0.834}{0.001}$ &
        $\ms{0.824}{0.001}$ &
        $\picpbest{0.823}{0.001}$ \\

        ~ & MPIW &
        $\ms{0.071}{0.000}$ &
        $\ms{0.066}{0.000}$ &
        $\ms{0.070}{0.000}$ &
        $\ms{0.055}{0.000}$ &
        $\ms{0.106}{0.001}$ &
        $\ms{0.095}{0.001}$ &
        $\ms{0.111}{0.000}$ &
        $\ms{0.080}{0.000}$ &
        $\ms{0.064}{0.000}$ &
        $\mpiwbest{0.060}{0.000}$ &
        $\ms{0.083}{0.000}$ \\


        ~ & MAE &
        $\ms{0.022}{0.000}$ &
        $\ms{0.022}{0.000}$ &
        $\ms{0.022}{0.000}$ &
        $\ms{0.030}{0.000}$ &
        $\maebest{0.032}{0.000}$ &
        $\ms{0.034}{0.000}$ &
        $\ms{0.036}{0.000}$ &
        $\ms{0.026}{0.000}$ &
        $\maebest{0.020}{0.000}$ &
        $\maebest{0.021}{0.000}$ &
        $\ms{0.034}{0.000}$ \\

        ~ & RMSE &
        $\ms{0.038}{0.000}$ &
        $\ms{0.037}{0.000}$ &
        $\ms{0.038}{0.000}$ &
        $\ms{0.036}{0.000}$ &
        $\ms{0.047}{0.000}$ &
        $\ms{0.048}{0.000}$ &
        $\ms{0.051}{0.000}$ &
        $\ms{0.042}{0.000}$ &
        $\rmsebest{0.036}{0.000}$ &
        $\ms{0.037}{0.000}$ &
        $\ms{0.048}{0.000}$ \\ \midrule

        Engression & CRPS &
        $\ms{0.199}{0.001}$ &
        $\crpsbest{0.101}{0.001}$ &
        $\ms{0.181}{0.001}$ &
        $\ms{0.146}{0.000}$ &
        $\crpsbest{0.413}{0.012}$ &
        $\crpsbest{0.186}{0.007}$ &
        $\ms{0.247}{0.010}$ &
        $\crpsbest{0.107}{0.001}$ &
        $\crpsbest{0.073}{0.000}$ &
        $\crpsbest{0.072}{0.001}$ &
        $\ms{0.448}{0.017}$ \\

        ~ & PICP &
        $\ms{0.448}{0.010}$ &
        $\ms{0.820}{0.004}$ &
        $\ms{0.735}{0.007}$ &
        $\ms{0.089}{0.004}$ &
        $\ms{0.758}{0.022}$ &
        $\picpbest{0.949}{0.003}$ &
        $\ms{0.859}{0.005}$ &
        $\ms{0.608}{0.006}$ &
        $\ms{0.843}{0.008}$ &
        $\ms{0.771}{0.005}$ &
        $\ms{0.441}{0.064}$ \\

        ~ & MPIW &
        $\mpiwbest{0.031}{0.001}$ &
        $\ms{0.046}{0.001}$ &
        $\mpiwbest{0.044}{0.001}$ &
        $\ms{0.089}{0.004}$ &
        $\ms{0.086}{0.001}$ &
        $\ms{0.090}{0.002}$ &
        $\mpiwbest{0.073}{0.003}$ &
        $\mpiwbest{0.029}{0.000}$ &
        $\mpiwbest{0.033}{0.001}$ &
        $\mpiwbest{0.030}{0.000}$ &
        $\ms{0.039}{0.003}$ \\


        ~ & MAE &
        $\ms{0.019}{0.000}$ &
        $\ms{0.015}{0.000}$ &
        $\ms{0.018}{0.000}$ &
        $\ms{0.033}{0.000}$ &
        $\ms{0.035}{0.000}$ &
        $\maebest{0.010}{0.000}$ &
        $\ms{0.019}{0.000}$ &
        $\ms{0.029}{0.000}$ &
        $\ms{0.031}{0.000}$ &
        $\ms{0.017}{0.000}$ &
        $\ms{0.031}{0.000}$ \\

        ~ & RMSE &
        $\ms{0.034}{0.000}$ &
        $\ms{0.028}{0.000}$ &
        $\ms{0.033}{0.000}$ &
        $\ms{0.033}{0.000}$ &
        $\ms{0.046}{0.001}$ &
        $\rmsebest{0.032}{0.001}$ &
        $\ms{0.034}{0.000}$ &
        $\rmsebest{0.031}{0.000}$ &
        $\rmsebest{0.024}{0.000}$ &
        $\rmsebest{0.025}{0.000}$ &
        $\ms{0.043}{0.001}$ \\

    \bottomrule
    \end{tabular}
    \end{adjustbox}
    \label{table:traffic_results}
\end{table*}

\clearpage

\bibliography{aaai2027}

\end{document}